\documentclass{clv3}
\usepackage{float}
\usepackage{times}
\usepackage{latexsym}
\usepackage{multirow}
\usepackage{graphicx}
\usepackage{amsmath}
\usepackage{comment}

\usepackage{url}

\usepackage{hyperref}
\usepackage{xcolor}
\definecolor{darkblue}{rgb}{0, 0, 0.5}
\hypersetup{colorlinks=true,citecolor=darkblue, linkcolor=darkblue, urlcolor=darkblue}
\usepackage{flushend}

\issue{.}{-}{2019}

\runningtitle{From bilingual to multilingual neural machine translation}

\runningauthor{Carlos Escolano, Marta R. Costa-juss\`a and Jos\'e A. R. Fonollosa}

\begin{document}

\title{From bilingual to multilingual Neural-based Machine Translation by Incremental Training}

\maketitle

\large{Carlos Escolano, Marta R. Costa-juss\`a, Jos\'e A. R. Fonollosa

\noindent{TALP Research Center, Universitat Polit\`ecnica de Catalunya, Barcelona}

\noindent{\{carlos.escolano,marta.ruiz,jose.fonollosa\}@upc.edu}}

\vspace{2cm}




\begin{abstract}
A common intermediate language representation in neural machine translation can be used to extend bilingual to multilingual systems by incremental training. In this paper, we propose a new architecture based on introducing an interlingual loss as an additional training objective. 
By adding and forcing this interlingual loss, we are able to train multiple encoders and decoders for each language, sharing a common intermediate representation. 

Translation results on the low-resourced tasks (Turkish-English and Kazakh-English tasks, from the popular Workshop on Machine Translation benchmark)  show the following BLEU improvements up to 2.8. However, results on a larger dataset (Russian-English and Kazakh-English, from the same baselines) show BLEU loses if the same amount. 
While our system is only providing improvements for the low-resourced tasks in terms of translation quality, our system is capable of quickly deploying new language pairs without retraining the rest of the system, which may be a game changer in some situations (i.e. in a disaster crisis where international help is required towards a small region or to develop some translation system for a client).  

Precisely, what is most relevant from our architecture is that it is capable of: (1) reducing the number of production systems, with respect to the number of languages, from quadratic to linear (2) incrementally adding a new language in the system without retraining languages previously there and (3) allowing for translations from the new language to all the others present in the system.


\end{abstract}

\section{Introduction}

Machine translation -in a highly multilingual environment- poses several challenges, as the number of possible combinations of translation directions grows quadratically. Among those challenges are the acquisition and curation of parallel data and the allocation of hardware resources for training and inference purposes. This situation becomes even worse given that the translation quality depends strongly on the amount of available training data when willing to offer translation for a language pair where there is little or no parallel data available.

Neural Machine Translation (neural machine translation) \cite{cho:2014,sutskever:2014} 
has arisen as a completely new paradigm for machine translation outperforming previous statistical approaches \cite{koehn:2003} in most of the tasks. One clear exception is low-resourced tasks \cite{koehn:2017}, where statistical machine translation still can outperform or be competitive with neural machine translation \cite{artetxe:2018,lample:2018}. 

Among others, one clear advantage of neural machine translation is that it opens new challenges in machine translation like multimodal machine translation \cite{elliott:2017:WMT} 
 or the finding of a common intermediate representation that allows training single encoders and decoders for each language reducing the number of translation systems from a quadratic dependency on languages to linear. As we will show in section \ref{section2}, there are different approaches that have used the idea of a common intermediate language in neural machine translation. However, recent research in this topic has been mainly on evaluating if the neural machine translation architecture of encoder-decoder with recurrent neural networks (RNNs), with or without attention mechanisms, is able to reach a universal language while training multiple languages \cite{johnson:2016,schwenk:2017}. These approaches can be further explored into unsupervised machine translation where the system learns to translate between languages without parallel data just by enforcing the generation and representation of the tokens to be similar \cite{artetxe:2017,lample:2017}.
All these architectures share parameters between languages and/or require all languages to be trained at the same time. This forces the system to be retrained in order to add new languages to the system and share the same representation. 
 
Differently, in this paper, we specifically pursue training a common intermediate representation for the benefit of incremental multilingual translation. Incremental translation is of high importance in situations like a natural disaster occurs like hurricane, tsunami or earthquake and humanitarian help is required from all over the world towards the place of the disaster. In situations where the place of the disaster has a minority language for which there are little speakers and/or non-machine translation systems available, a quick deployed machine translation system has been shown to be of vital need \cite{Lewis2010HaitianCH}. Other than these, companies may be interested in deploying new machine translation systems for clients without retraining the other languages that they have. The incremental multilingual translation that we are presenting is supported on a common intermediate representation which implies that either encoders or decoders of new languages have to be trained with only one encoder or decoder from the system, while keeping this last one frozen. 
Therefore, the proposed approach differs from the previous cases in which the intermediate representation is an end by itself (e.g., \cite{conneau:2017,conneau:2018}). Our proposed architecture combines variational autoencoders and the state-of-the-art Transformer architecture \cite{vaswani2017attention}. In our case, we are using specifically the vector quantization strategy as type of variational autoencoder and the main motivation for this is matching several close sentences in the same quantization step. However, this vector quantization strategy results in worse translation quality results. Another novelty of our architecture is that, in the optimisation process, we are adding a loss term. This loss term is the correlation between intermediate representations from different languages. Like this, we are forcing the system to learn the intermediate representation while training multiple translation systems. One of the challenges at this point is to find a suitable intermediate representation distance function. In order to address this, we propose and evaluate different distance measures. Results on the Workshop on Machine Translation (WMT) benchmark on low-resourced tasks (English, Turkish and Kazakh)\footnote{http://www.statmt.org/wmt19/} show that our architecture (with the new loss term based on correlation distance and without variational autoencoders) produces competitive translations and improves over the baseline system while benefiting from an easily and inexpensively way of extending to new languages. Regarding adding new languages, we show that our architecture is capable of scaling to new source languages without requiring re-training all languages in the system.

\paragraph{Contribution} This paper proposes a proof of concept of a new multilingual neural machine translation approach. The current approach is based on joint training without parameter sharing by enforcing a compatible representation between the jointly trained languages and using multitask learning. 
We are proposing to effectively use the correlation as the distance between intermediate representations. This approach is shown to offer a scalable strategy to new languages without retraining any of the previous languages in the system and enabling zero-shot translation. Our architecture while being much more flexible and allowing for quickly deploying new languages is also able to improve BLEU results over the baseline systems.

\paragraph{Organisation} The rest of the paper is organised as follows. Section \ref{section2} reports the most closely related work on the topic. Next Section \ref{background} explains the necessary background to make the manuscript self-contained. Section \ref{architecture} details the architecture proposed in this study, both the joint training and how to scale to new languages. Section \ref{experimental} describes the data and implementation used in the experiments, as well as it reports the translation results. Section \ref{visualization} provides an analysis of how the system is able to recover sentence from the intermediate representation and also an insightful visualization of the intermediate representation. Finally, section \ref{conclusions} depicts the most relevant conclusions of this study.

\section{Related Work}
\label{section2}


In this work, we are focusing on training a common language representation with deep learning techniques. The objective is to train an intermediate representation that allows using independent encoders and decoders for each language. In this scenario, translation systems in a highly multilingual environment get reduced from quadratic to linear and also, translation is available for language pairs that have not been explicitly trained. 
Following a similar objective or methodology, related works include the following ones.

\paragraph{Shared Encoders/Decoders} Johnson et al. (2016) \nocite{johnson:2016} feed a single encoder and decoder with multiple input and output languages. With this approach, authors show that zero-shot learning is possible. Authors show by means of visualizing similar sentences in different languages that there is some hint that these appear somehow close in the common representation. More recently, Arivazhagan et al (2019) \nocite{DBLP:journals/corr/abs-1903-07091} propose auxiliary losses on the neural machine translation encoder
that impose representational invariance across languages, which is shown to improve zero-shot translation.

\paragraph{Dedicated Encoder/Decoder} These approaches vary from many encoders to one decoder (many-to-one) \cite{zoph:2016}, one encoder to many decoders (one-to-many) \cite{dong:2016} and, finally, one encoder to one decoder (one-to-one), which we are focusing on because they are closest to our approach. Firat et al. 2016 \nocite{firat:2017} propose to extend the classical recurrent neural machine translation bilingual architecture \cite{bahdanau:2015} to multilingual by designing a single encoder and decoder for each language with a shared attention-based mechanism. Schwenk et al. (2017) \nocite{schwenk:2017} and Espana-Bonet et al. (2017) \nocite{espana:2017} evaluate how a recurrent neural machine translation architecture without attention is able to generate a common representation between languages. Authors use the inner product or cosine distance to evaluate the distance between sentence representations. Recently, Lu et al., (2018) \nocite{lu:2017} train single encoders and decoders for each language generating interlingual embeddings which are agnostic to the input and output languages. 

\paragraph{Other related architectures} While unsupervised machine translation \cite{lample:2017,artetxe:2017} is not directly pursuing a common intermediate representation, but it is somehow related to our approach. Artetxe et al. (2017) and Lample et al. (2017) 
propose a translation system that is able to translate trained only on a monolingual corpus. The architecture is basically a shared encoder with pre-trained embeddings and two decoders (one of them includes an autoencoder). On the other hand, our work is also related to recent works on sentence representations \cite{conneau:2017,conneau:2018,eriguchi:2018} and taking advantage of multitask learning \cite{dong2015multi}. However, the main difference is that these works aim at extending representations to other natural language processing tasks, while we are aiming at finding the most suitable representation to make interlingual machine translation feasible. While interesting for further research, it is out-of-scope of this study the evaluation and adaptation of this intermediate representation to multiple tasks.

\section{Background}
\label{background}

In this section, we report the techniques that are used for the development of the proposed architecture in this paper: variational autoencoders \cite{rumelhart1985learning}, decomposed vector quantization \cite{oord:2017} and Transformer \cite{vaswani2017attention}. 

\subsection{Variational Autoencoders}

Autoencoders consist of a generative model that is able to generate its own input. This is useful to train an intermediate representation, which can be later employed as a feature for another task or even as a dimensionality reduction technique. This is the case of traditional autoencoders that learn to produce an intermediate representation for an existing example. Variational autoencoders \cite{rumelhart1985learning,kingma:2013,zhang-EtAl:2016:EMNLP20162} present a different approach in which the objective is to learn the parameters of a probability distribution that characterizes the intermediate representation. This allows to sample new synthetic instances from the distribution and generate them using the decoder part of the network.

\subsection{Decomposed Vector Quantization}

One of the strategies to create variational autoencoders is vector quantization \cite{oord:2017}. This consists of the addition of a table of dimension $K \cdot D$ where $K$ is the number of possible representations and $D$ the dimension or set of dimensions of each of the representations. The closest vector to the output of the encoder of the network is fed to the decoder as a discrete latent representation that would be employed to reconstruct the input.

As proposed in \cite{kaiser2018fast} this approach may produce a vector quantization in which only a small part of the vectors is employed. To solve this, \textit{decomposed vector quantization} uses a set of $n$ tables in which each table is used to represent a portion of the representation that would be later concatenated and fed to the decoder. This approach presents the advantage that by using $n$ $K \cdot \frac{D}{n}$ tables and optimizing the same number of parameters, $K^n$ possible vectors of dimension $D$ can be generated.




\subsection{Transformer}

The current state-of-the-art architecture for neural machine translation is the Transformer \cite{vaswani2017attention} and it uses multiple self-attention and feed-forward layers, which allow to deal with sequences by inputting the whole sequence at once and using self-attention for attending to the relevant parts of the sequence and solving correference issues. Apart from improving the results of the previous sequence-to-sequence systems \cite{vaswani2017attention}, the Transformer can run in parallel. Among other tricks and particularities, this architecture requires summing positional embeddings, as previously proposed in \cite{gehring-etal-2017-convolutional}, to word embeddings for explicitly encoding the relative positions of tokens, because being a non-recurrent architecure it does not have an inherent notion of sequentiality.

\section{Model Architecture}
\label{architecture}

In this section, we report the details of our proposed architecture. We describe the joint training and how we are scaling to new languages.

\subsection{Definitions}

Before explaining our proposed model we introduce the annotation that will be assumed hereinafter. Languages will be referred to as capital letters $X,Y,Z$ while sentences will be referred in lower case $x,y,z$ given that $x\in X$, $y \in Y$, and $z \in Z$.  Then, sentence $i$ in the corpus data is referred to as $x_i,y_i,z_i$.

We consider as an encoder ($e_x, e_y,e_z$) the layers of the network that given an input sentence produce a sentence representation in a space.  Analogously, a decoder ($d_x,d_y,d_z$) is the layers of the network that given the sentence representation of the source sentence is able to produce the tokens of the target sentence. Encoders and decoders will be always considered as independent modules that can be arranged and combined individually as no parameter is shared between them. Each language and module has its own weights independent from all the others present in the system.

\subsection{Joint Training}

Given two languages $X$ and $Y$, our objective is to train independent encoders and decoders for each language, $e_x, d_x$ and $e_y, d_y$ that produce compatible sentence representations. For instance, given a sentence $x$ in language $X$, we can obtain a representation $r_x$ from that the encoder $e_x$ that can be used to either generate a sentence reconstruction using decoder $d_x$ or a translation using decoder $d_y$. 
With this objective in mind, we propose a training schedule that combines two tasks (auto-encoding and translation) and the two translation directions simultaneously by optimizing the following loss:

{\small
\begin{equation}
L = L_{XX} + L_{YY} + L_{XY} + L_{YX} + d
\end{equation}
}
where $L_{XX}$ and $L_{YY}$ correspond to the reconstruction losses of both language $X$ and $Y$ (defined as the cross-entropy of the generated tokens and the source sentence for each language); $L_{XY}$ and $L_{YX}$ correspond to the translation terms of the loss measuring token generation of each decoder given a representation generation by the other language decoder (using the cross-entropy between the generated tokens and the translation reference); and $d$ corresponds to the distance metric between the representation computed by the encoders. This last term forces the representations to be similar without sharing parameters while providing a measure of similarity between the generated spaces.
We have tested different distance metrics such as L1, L2 or the discriminator addition (that tried to predict from which language the representation was generated). For all these alternatives, we experienced a space collapse in which all sentences tend to be located at the same spatial region. This closeness between the sentences of the same languages makes them non-informative for decoding. As a consequence, the decoder performs as a language model, producing an output only based on the information provided by the previously decoded tokens. To prevent this collapse, we propose a less restrictive measure based on correlation distance \cite{chandar:2015} computed as in equations \ref{eq:corr_dist} and \ref{eq:corr_dist2}. The rationale behind this loss is maximizing the correlation between the representations produced by each language while not enforcing the distance over the individual values of the representations.

\begin{equation} \label{eq:corr_dist}
 d = 1 - c(h(X),h(Y))  
\end{equation} 
 
\begin{equation}  
\label{eq:corr_dist2}
 c(h(X),h(Y)) =   
 \frac{\sum^{n}_{i=1}(h(x_{i} - \overline{h(X)})(h(y_{i} - \overline{h(Y)})}{\sqrt[]{\sum^{n}_{i}(h(x_{i})-\overline{h(X)})^{2}\sum^{n}_{i}(h(y_{i}) -\overline{h(Y)})^{2}}}  
\end{equation} 
%

where $X$ and $Y$ correspond to the data sources we are trying to represent; $h(x_{i})$ and $h(y_{i})$ correspond to the intermediate representations learned by the network for a given $i$ observation; and $\overline{h(X)}$ and $\overline{h(Y)}$ are, for a given batch, the intermediate representation mean of $X$ and $Y$, respectively.

\begin{figure}[h]
\begin{center}
\includegraphics[scale=0.35]{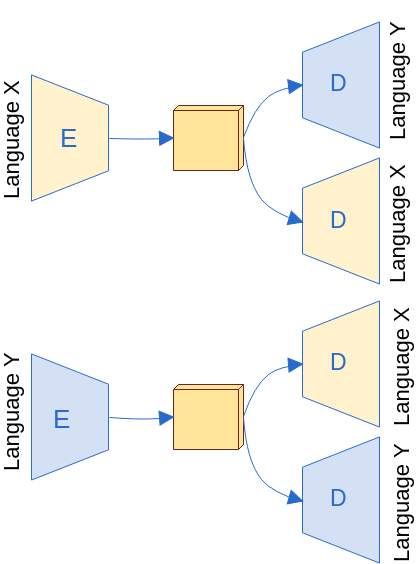}
\caption{Joint training scheme. Only one language X and language Y are trained, repeated in the figure to show all possibilities.}
\label{img:joint-training}
\end{center}
\vspace{-0.75cm}
\end{figure}

Figure \ref{img:joint-training} shows the different task and directions that the system has been trained to perform. Each decoder is able to process the representation produced by each encoder to either translate or reconstruct the source language sentence.

\subsection{Scaling to new languages}
\label{sec:scaling}

Given the jointly trained model between languages $X$ and $Y$, the following step is to add new languages in order to use our architecture as a multilingual system. Since parameters are not shared between the independent encoders and decoders, our architecture enables to add new languages without the need to retrain the current languages in the system. 
Let's say we want to add language $Z$. To do so, we require to have parallel data between $Z$ and any language in the system. So, assuming that we have trained $X$ and $Y$, we need to have either $Z-X$ or $Z-Y$ parallel data. For illustration, let's fix that we have $Z-X$ parallel data. Then, we can set up a new bilingual system with language $Z$ as source and language $X$ as target. To ensure that the representation produced by this new pair is compatible with the previously jointly trained system, we use the previous $X$ decoder ($d_x$) as the decoder of the new $ZX$ system and we freeze it. During training, we optimize the cross-entropy between the generated tokens and the language $X$ reference data but only updating the layers belonging to the language $Z$ encoder ($e_z$). Doing this, we train $e_z$ not only to produce good quality translations but also to produce similar representations to the already trained languages. 


\begin{figure}[h]
\begin{center}
\includegraphics[scale=0.35]{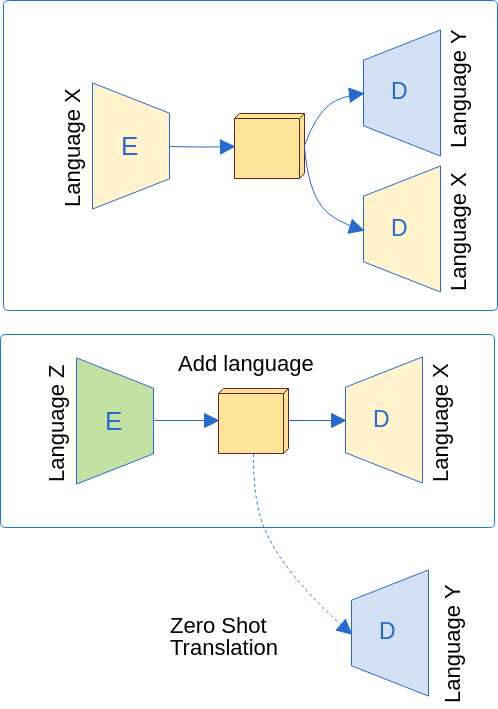}
\caption{Language addition and zero shot training scheme}
\label{img:scalability}
\end{center}
\vspace{-0.75cm}
\end{figure}

Our training schedule enforces the generation of a compatible representation, which means that the newly trained encoder $e_z$ can be used as input of the decoder $d_y$ from the jointly trained system to produce zero-shot $Z$ to $Y$ translations. See Figure \ref{img:scalability} for illustration.
The fact that the system enables zero-shot translation shows that the representations produced by our training schedule contain useful information and that this can be preserved and shared to new languages just by enforcing the new modules to train with the previous one, without any modification of the architecture.

A current limitation is the need to use the same vocabulary for the shared language ($X$) in both training steps. The use of subwords \cite{sennrich2015neural} mitigates the impact of this constraint.


\section{Experimental framework}
\label{experimental}

In this section, we provide details about the data and implementation for the experiments. Additionally, we report the translation results. Results are presented in terms of BLEU \cite{bleu} which is the standard automatic measure in machine translation. 

\subsection{Data}

For experiments, we use the Turkish-English parallel data from \textit{setimes2} \cite{tiedemann:2009} which is used in WMT 2017 \footnote{http://www.statmachine translation.org/WMT17/} and the Kazakh-English parallel data from the news domain which
is used in WMT 2019 \footnote{http://www.statmt.org/WMT19/}. The training set for Turkish-English is around 200,000 parallel sentences and for the Kazakh-English is around 100,000 parallel sentences. As development and test set we used \textit{newsdev2016} and \textit{newstest2016}, respectively, for Turkish-English and \textit{newsdev2019} was split into development and test set for Kazakh-English experiments. Additionally, we extracted the Kazakh-Turkish test set from the
OPUS database \cite{opus} to evaluate the zero-shot translation.
For experiments with larger datasets we used data between Russian-English used in WMT 2019 \footnote{http://www.statmt.org/WMT19/} and between Russian-Kazakh. The validation and test sets from Russian-English were extracted from the \textit{Yandex} corpus. Validation set for Russian-Kazakh was extracted from news-commentary-v14. Finally, and only for visualization and analysis purposes, we extracted 381 sentences which are multi-way parallel in Turkish-Kazakh-English. These sentences are also extracted from the OPUS database\footnote{The datasets that we prepared for Kazakh-Turkish and Turkish-Kazakh-English, which are the only ones that not belong to a benchmark, are freely available under request.} For the latter, we download TK-EN and KK-EN and did the matching of the English sentences that were identical. Detailed statistics of the corpus are shown in Table \ref{tab:corpora}.

\begin{table*}[h]
\caption{Size of the parallel corpora}
\label{tab:corpora}
\scriptsize
\centering
\medskip
   \begin{tabular}{l| rrrrr|r}
Language Pair &  {\bf Corpus} & Language  & Set & {Segments} & { Words} & {Vocab (BPE)}\\ \hline
&   & TK   &  & &  $ 5513 \cdot 10^3$ & \\ 
& \raisebox{1.5 ex}[0pt]{SETIMES}
& EN  & \raisebox{1.5 ex}[0pt]{Training} & \raisebox{1.5 ex}[0pt]{$~ 200 \cdot 10^3$}  & $~ 5467 \cdot 10^3$ &  \\
\cline{2-6}
&& TK   &  & &  $ 26,7 \cdot 10^3$ & \raisebox{0 ex}[0pt]{$ 16 \cdot 10^3$}  \\ 
\raisebox{1.5 ex}[0pt]{TK-EN} &  \raisebox{1.5 ex}[0pt]{newsdev 2017}
& EN  & \raisebox{1.5 ex}[0pt]{Validation} & \raisebox{1.5 ex}[0pt]{$~1,001 \cdot 10^3$}  & $~ 28,4 \cdot 10^3$ & \raisebox{0.0 ex}[0pt]{$ 16 \cdot 10^3$} \\
\cline{2-6}
&& TK   &  & &  $ 84,6 \cdot 10^3$ &  \\ 
&  \raisebox{1.5 ex}[0pt]{newstest 2017}
& EN  & \raisebox{1.5 ex}[0pt]{Test} & \raisebox{1.5 ex}[0pt]{$~3 \cdot 10^3$}  & $~ 86,6 \cdot 10^3$ &  \\
\hline
&& KK   & & &  $ 2219 \cdot 10^3$ &  \\ 
& \raisebox{1.5 ex}[0pt]{News Commentary v14}
& EN &\raisebox{1.5 ex}[0pt]{Training}  & \raisebox{1.5 ex}[0pt]{$~100 \cdot 10^3$} & $~ 2216 \cdot 10^3$ &  \\ 
\cline{2-6}
&& KK   &  & &  $35,5 \cdot 10^3$ &  \\ 
\raisebox{1.5 ex}[0pt]{KK-EN} &  \raisebox{1.5 ex}[0pt]{newsdev 2019}
& EN  & \raisebox{1.5 ex}[0pt]{Validation} & \raisebox{1.5 ex}[0pt]{$~1,033 \cdot 10^3$}  & $~36,7 \cdot 10^3$ &  \\
\cline{2-6}
&& KK   &  & &  $ 36,7 \cdot 10^3$ & \raisebox{0.0 ex}[0pt]{$7,36 \cdot 10^3$}  \\ 
&  \raisebox{1.5 ex}[0pt]{newstest 2019}
& EN  & \raisebox{1.5 ex}[0pt]{Test} & \raisebox{1.5 ex}[0pt]{$~1,033 \cdot 10^3$}  & $~38 \cdot 10^3$ &  \raisebox{0.0 ex}[0pt]{$ 8,99 \cdot 10^3$}\\
\cline{1-6}
&& KK   &  & &  $22,5 \cdot 10^3$ & \\
\raisebox{1.5 ex}[0pt]{KK-TK} &  \raisebox{1.5 ex}[0pt]{OpenSubtitles}
& TK  & \raisebox{1.5 ex}[0pt]{Test} & \raisebox{1.5 ex}[0pt]{$~2,594 \cdot 10^3$}  & $~26,3 \cdot 10^3$ &   \\ \cline{1-6}
&& KK   &  & &  $  1,435 \cdot 10^3$ &  \\ 
KK-TK-EN & Tatoeba & TK  & {Test} & $381$  & $~1,369 \cdot 10^3$ &  \\ 
& & EN  &  &   & $~ 1,827 \cdot 10^3$ &  \\ \hline\hline
&& RU   &  & &  $180 \cdot 10^3$ &  \\ 
&  \raisebox{1.5 ex}[0pt]{Yandex Corpus + ParaCrawl}
& EN  & \raisebox{1.5 ex}[0pt]{Training} & \raisebox{1.5 ex}[0pt]{$~6,066 \cdot 10^3$}  & $~ 311 \cdot 10^6$ & \\
\cline{2-6}
&& RU   &  & &  $117,3 \cdot 10^3$ & $31,8 \cdot 10^3$ \\ 
\raisebox{1.5 ex}[0pt]{RU-EN} &  \raisebox{1.5 ex}[0pt]{Newsdev 2019}
& EN  & \raisebox{1.5 ex}[0pt]{Validation} & \raisebox{1.5 ex}[0pt]{$~4, \cdot 10^3$}  &$~204,1 \cdot 10^3$  & $ 32 \cdot 10^3$   \\
\cline{2-6}
&& RU   &  & &  $ 29,8 \cdot 10^3$ &  \\ 
& \raisebox{1.5 ex}[0pt]{Newstest 2019}
& EN  & \raisebox{1.5 ex}[0pt]{Test} & \raisebox{1.5 ex}[0pt]{$~1 \cdot 10^3$}  & $~51,2 \cdot 10^3$ &  \\
\hline
&& KK   & & &  $ 284,9 \cdot 10^6$ & \\ 
& \raisebox{1.5 ex}[0pt]{WMT19 Crawled Corpus}
& RU &\raisebox{1.5 ex}[0pt]{Training}  & \raisebox{1.5 ex}[0pt]{$~4,18 \cdot 10^6$} & $~ 109,1 \cdot 10^6$ &  \\ 
\cline{2-6}
&& KK   &  & &  $266,6 \cdot 10^3$ & $ 24,2 \cdot 10^3$  \\ 
\raisebox{1.5 ex}[0pt]{KK-RU} &  \raisebox{1.5 ex}[0pt]{Ext. WMT19 Crawled Corpus}
& RU  & \raisebox{1.5 ex}[0pt]{Validation} & \raisebox{1.5 ex}[0pt]{$~4 \cdot 10^3$}  & $~ 102,2 \cdot 10^3$ & $ 34,6 \cdot 10^3$ \\
\cline{2-6}
&& KK   &  & &  $66,9 \cdot 10^3$ &  \\ 
& \raisebox{1.5 ex}[0pt]{Ext. WMT19 Crawled Corpus}
& RU  & \raisebox{1.5 ex}[0pt]{Test} & \raisebox{1.5 ex}[0pt]{$~1  \cdot 10^3$}  & $~ 25,4 \cdot 10^3$ &  \\
\hline
\end{tabular}
\end{table*}


Preprocessing consisted of a pipeline of punctuation normalization, tokenization, corpus filtering of longer sentences than 80 words and true-casing. These steps were performed using the scripts available from Moses tools \cite{koehn2007moses}. In the experiments using subwords, preprocessed data is tokenized using Byte Pair Encoding (BPE) \cite{sennrich:2015}.

\subsection{Implementation}

We used the Transformer implementation provided by Fairseq\footnote{Release v0.6.0 available at https://github.com/pytorch/fairseq}. Parameters varied depending on the configuration of the experiment. For experiments within configuration 1 (conf1), we used 6 blocks of multihead attention of 8 heads each,  embedding/hidden dimensionality of 128 and fixed learning rate of 0.001 and vocabulary size of 12,000 words. For experiments within configuration 2 (conf2), we used 6 attention blocks with 4 heads, embedding/hidden dimensionality of 512, and a fixed learning rate of 0.001 and vocabulary size of 16000 BPE tokens. For all cases, we used Adam \cite{kingma2014adam} as the optimizer. The joint training was performed in two Nvidia Titan X GPUs with 12 GB of RAM while the addition of languages used one Titan X GPU. As for stopping criterion, systems trained until non-improvement was seen on the validation set.



\subsection{Translation quality for the Joint Training}
\label{sec:transqual}

Table \ref{results} shows BLEU results for two experimental configurations (1 \& 2) as reported in above and for each translation direction from English-to-Turkish (EN-TK) and from Turkish-to-English (TK-EN). Within the first configuration we shows results for variations on the proposed architecture (JointTrain) which include 
both non-variational and variational (dqv) with the same hyperparameters of the baseline system and a comparison of two distance losses. 

\paragraph{Variational vs non-variational} When using the decomposed vector quantization, the performance gets worse in both directions and the loss is higher in EN-TK. When contrasting the impact of the decomposed vector quantization in our proposed architecture, we see that the performance of non-variational architecture is also higher than the decomposed vector quantization (dvq) using any type of distance. However, the loss, when using the correlation distance, is higher in the direction of TK-EN than in the opposite. 

\paragraph{Correlation vs Maximum distance loss} In regard to the distance loss, the correlation distance clearly provides better translation results, by approximately 1.5 \textit{BLEU} in both directions when using the non-variational architecture. The improvement of the correlation distance compared to the maximum distance is even higher when using the variational architecture. 

\begin{table}[!h]
\centering
\caption{BLEU results for the different experimental configuration (1 \& 2) and alternative architectures: the Transformer system (baseline) and different configurations of our architecture, our Joint Training (JoinTrain) with and without decomposed vector quantization (dvq), and correlation distance (corr) and maximum of difference (max). Best results in bold.}
\label{results}
\begin{tabular}{l|l|l|l}
\cline{2-4}
                                         & Config & \textbf{EN-TK} & \textbf{TK-EN} \\ \hline
\multicolumn{1}{l|}{Baseline}        & 1 & 8.32           & 12.03          \\ \hline
\multicolumn{1}{l|}{Baseline dvq}    & 1 & 2.89           &  8.14          \\ \hline
\multicolumn{1}{l|}{JoinTrain + corr}       & 1 & 8.11           & 12.00          \\ \hline
\multicolumn{1}{l|}{JoinTrain + max}        & 1 & 6.19           & 10.38          \\ \hline
\multicolumn{1}{l|}{JoinTrain + dvq + corr} & 1 & 7.45           &  7.56          \\ \hline
\multicolumn{1}{l|}{JoinTrain + dvq + max}  & 1 & 2.40           &  5.24          \\ \hline\hline
\multicolumn{1}{l|}{Baseline} & 2 & 11.85  &  14.31\\ \hline
\multicolumn{1}{l|}{JoinTrain+corr} & 2 & \textbf{12.56}      & \textbf{14.82} \\ \hline
\multicolumn{1}{l|}{JoinTrain+corr+no-autoencoders}  & 2 & 12.16  & 14.21 \\ \hline
\end{tabular}
\end{table}

Within this best configuration, our best proposed architecture, which is non-variational autoencoder with correlation distance (JoinTrain + corr) shows similar performance compared to the baseline system (Transformer).


Up to here, experiments were performed using words. At this point, we are changing from we employ \textit{subword-neural machine translation} \cite{sennrich:2015} which is standard tokenization of words. Our configuration follows the standard set-up with 16000 operations and shared vocabulary between both languages. Second part of Table \ref{results} shows the performance of our best architecture from configuration 1 against the baseline system with the second experimental configuration (using \textit{BPE} as tokenization and a larger word embedding of 512).



We achieve gains of +0.5 BLEU from English-to-Turkish and +0.7 BLEU from Turkish-to-English over the corresponding baseline. At this point, we found interesting training our architecture without autoencoders, and we see that it improves over the baseline system but not over our complete original proposed joint training. Therefore, training with autoencoders helps improving translation performance.

Note that the performance of both the baseline and our architecture in this second configuration is higher than the best system results from WMT 2017 \cite{mercedes:2017}. We are comparing to the case of using only parallel data, without adding back-translated monolingual data (which were 10.9 for EN-TK and 14.2 for TK-EN).



\subsection{Adding new languages and Zero-shot Translation}
\label{sec:adding}

At this point, we use the best configuration from the above experiments (configuration 2, using BPE and 512 word embedding dimension).
We add Kazakh as a new language to this system as proposed in section \ref{sec:scaling}. 
 Table \ref{bleu-bpe-add} shows that Kazakh-English performs +0.6 BLEU points over the baseline. The frozen English decoder previously trained using the Turkish-English parallel data may be responsible for the increase of performance.  
 




Finally, another relevant aspect of the proposed architecture is enabling zero-shot translation. To evaluate it, we compare the performance of Kazakh-Turkish compared to a pivot system based on the cascade. 
Such a system consists of translating from Kazakh to English and from English to Turkish. We can do this pivot either with the baseline system (standard Transformer) or with our JoinTrain architecture modules. Finally, we can use the ZeroShot strategy which consists in using Kazakh encoder and Turkish decoder. Results show that the zero-shot translation provides slightly lower quality than the pivot systems, but the pivot joint training improves over the pivot baseline. 


\begin{table}[h!]
\centering
\caption{New supervised language (KK-EN) comparing: the baseline architecture to our added language (AddLang). Zero-shot translation (KK-TK) provided by our architecture compared to a baseline which is a pivot system from KK-EN and EN-TK. Best results in bold.}
\label{bleu-bpe-add}
\begin{tabular}{l|l|l|l|l}
\hline
Config & System & \textbf{KK-EN}  & System & \textbf{KK-TK}\\ \hline
2&\multicolumn{1}{l|}{Baseline} & 3,82 & Pivot Baseline & 4,74\\ \hline
2&\multicolumn{1}{l|}{AddLang +corr} & \textbf{4,44}  &  Pivot JointTrain & \textbf{4.93} \\ \hline
&  &&  ZeroShot & 4,36\\ \hline
\end{tabular}
\end{table}

Results from Table \ref{bleu-bpe-add} may seem low, but note that our system in Kazakh-to-English gets comparable results to those systems that participated in the WMT 2019 evaluation that did not use monolingual data and backtranslate it\footnote{http://matrix.statmt.org/}.

\subsection{Larger Datasets}

One of the main advantages of shared architecture multilingual NMT systems\cite{johnson:2016} is that they allow by parameter sharing the flow of information between high-resourced languages and low-resourced languages. As a result, the performance of the later ones improved.

In this experiment, we want to confirm that our proposed architecture also presents this property, by adding a new resource language with a frozen model trained in a high-resourced language. To do so, we trained a Russian-English model and add Kazakh as low-resourced language. 

We added Kazakh to the already trained Russian decoder from the Russian-English trained system.

As the objective of this experiment is only to measure the impact of the additional data for low resources the vocabularies were shared between all three languages, including two different script systems, Latin for English and Cyrillic for Russian and Kazakh. Table \ref{results2} shows the results for the baselines and our architecture between Russian and English.

\begin{table}
\centering{
\caption{BLEU results for the larger dataset RU-EN.}
\label{results4}
\begin{tabular}{l|l|l|l}
\cline{2-4}
                                         & Config & \textbf{EN-RU} & \textbf{RU-EN} \\ \hline
\multicolumn{1}{l|}{Baseline} & 2 &  \textbf{32.33} & \textbf{35.09} \\ \hline
\multicolumn{1}{l|}{JoinTrain+corr} & 2 &  29.48     & 34.64 \\ \hline
\hline
\end{tabular}}
\end{table}

\begin{table}[h!]
\centering
\caption{Zero-shot translation (KK-EN) provided by our architecture compared to: a low-resource baseline (KK-EN), a pivot system from KK-RU and RU-EN both the baseline and our JointTrain architecture. Best results in bold.}
\label{table:low-resource}
\begin{tabular}{l|l|l}
\hline
Config & System & \textbf{KK-EN}\\ \hline
2  & Baseline & 2.32 \\ \hline
2& Pivot Baseline & \textbf{16.62} \\ \hline
2   &  Pivot JointTrain & 13.69\\ \hline
2&  ZeroShot & 5.58 \\ \hline
\end{tabular}
\end{table}


As the system benefits from more data available, we want to evaluate its performance in the case of training Kazakh with a larger dataset and perform zero-shot translation from Kazakh to English, so as mentioned, we are adding Kazakh to the system by training a Kazakh encoder with the already trained Russian decoder. Results in Table \ref{table:low-resource} show that this zero-shot approach outperforms both baseline and direct Kazakh to English addition, proving that having access to more data provides better models in this architecture. This idea holds even when to parallel data has been used only during the training and the translation is based only in the compatibility of the sentence representations.

Nota that both table \ref{results4} and \ref{results5} show that our proposed system has a lower performance than the baseline system when trained on this new larger dataset. 


\section{Encoders/Decoders Compatibility and Intermediate Representation Visualization}
\label{visualization}

Our main objective is creating an intermediate representation that can be understood
by the different modules trained in the system, where the modules are the
encoders and decoders of all languages involved in training. But similar representations
may not lead to compatible encoder/decoders. Also, different trainings
can produce representations with different mean distances in the representations
that can generate similar translation outputs. With this focus in mind, in this section we do further analysis of our architecture by looking at the compatibility between encoders and decoders and visualising the intermediate representation. The model used for the analysis within this section is the \textit{JoinTrain+corr} and configuration 2, which is the best performing model from Table \ref{results} for Turkish-English and Kazakh-English from Table \ref{bleu-bpe-add}.

\subsection{Encoders/Decoders Compatibility}


We propose the following analysis towards measuring the compatibility of our encoders and decoders. Given a parallel set of sentences in the languages in which the system has been trained, we can generate $e_{x}$ and $e_{y}$. Both encodings, coming from a parallel test, have the same number of vectors each of them of the same dimensionality. 

Our proposed analysis consists of inferring one of the decoders in the system ($d_x$ and $d_y$) using $e_{x}$ and $e_{y}$ as input. This generates two different outputs: an autoencoding output and a machine translation output. As we have parallel references for both languages we can measure the BLEU score \cite{bleu} of each of the results against the reference to measure how the models perform. 

\begin{figure}[h!]
\centering
\includegraphics[scale=0.25]{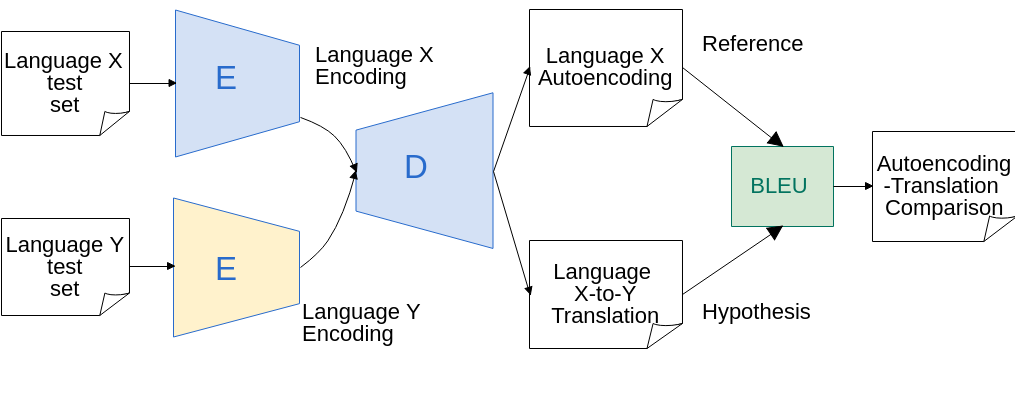}
\caption{Evaluation measure pipeline.}
\label{measure}
\end{figure}

Additionally, we can calculate a new BLEU score comparing the outputs of the autoencoding and the machine translation outputs. In the ideal case, encoders from two different languages have to produce the same representation for the same sentences. Therefore, the difference between the BLEU score obtained in the autoencoding output and the translation output shows how different are $e_{x}$ and $e_{y}$ representations in terms of how the decoder is able to generate accurate results from them. Our analysis consists in evaluating the BLEU score using the autoencoding output as a reference and the machine translation output as a hypothesis (this comparison will be refer to as A-T). Figure \ref{measure} shows the full pipeline of this procedure.

Table \ref{bleu-decoders} shows that the quality of the output of the decoder is quite better when the input comes from the encoder of the same language (autoencoder) than from another (machine translation). We also included the BLEU score between both autoencoder and translation outputs (A-T), which is the measure that we are proposing to evaluate the quality of our intermediate representation. Low BLEUs in \textit{A-T} indicates that we are still far from being able to decode from the common intermediate representation.

\begin{table}[]
\centering
\caption{Comparison of BLEU scores on the \textit{JointTrain+corr} architecture (on configuration 2) when performing as autoencoder and machine translation. The third column is the BLEU between autoencoder and translation outputs}
\label{bleu-decoders}
\begin{tabular}{|l|l|l|l|}
\hline
\textbf{Decoder} & \textbf{Autoencoder} & \textbf{machine translation} & \textbf{A-T} \\ \hline
EN               &         90.50            &       14.82             &     14.91   \\ \hline
TR               &         96.20            &       12.56             &      14.56   \\ \hline

\end{tabular}
\end{table}

\subsection{Visualization and Translation Examples}

As follows, we analyze the intermediate representations at the last attention block of the encoders, where we are forcing the similarity. In order to graphically show the presentation, we are using an in-house visualization tool \cite{demopaper} freely available\footnote{https://github.com/elorala/interlingua-visualization}. The tool trains a UMAP \cite{mcinnes2018umap-software} model combining the representations of languages that makes a dimensionality reduction of the sentence representations. 

\begin{figure}
\centering
        \includegraphics[width=0.7\textwidth]{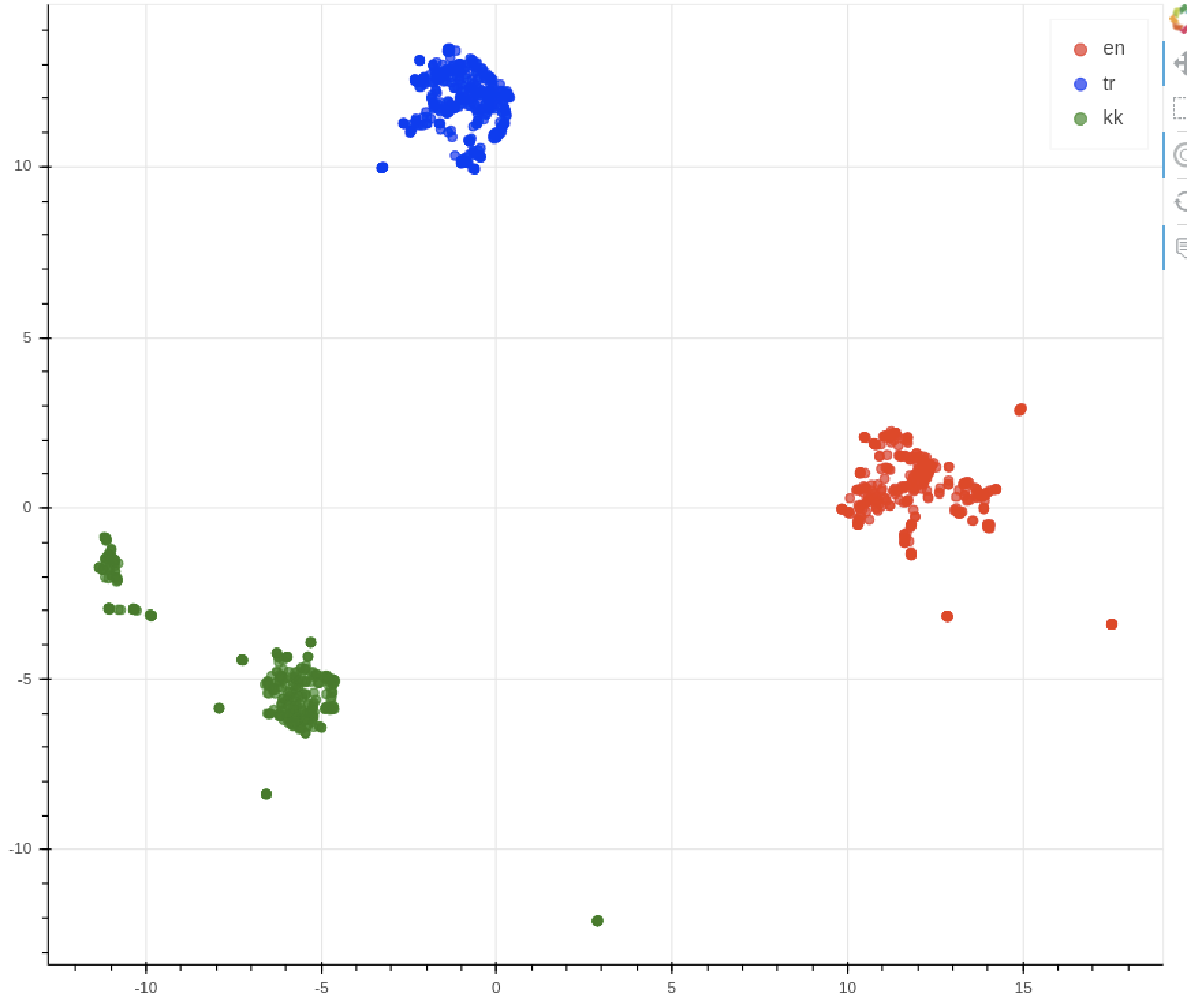}
    \label{img:vis-encoders}
    \caption{Encoder representations of the multi-way parallel 381 sentences. Turkish sentence representations (blue) compared to English sentence representations (red) and Kazakh sentences (green)}
\end{figure}

As follows, we used the 381 multi-way parallel sentences extracted specifically for this analysis (see statistics in Table \ref{tab:corpora}). 
Figure \ref{img:vis-encoders} shows the sentence representations created by their encoders. The separated clusters show that languages are not yet represented in the same space. Related work \cite{arivazhagan2018missing} shows similar results for the case of a multilingual system with shared encoder and decoder. While the system is able to produce compatible representations clear clusters can be observed for each language in the system. Plausible explanations for this difference may be the distance measure that we are using and/or the alignment of the source sentences. Some distance measures cause the representations to collapse in a small region of the space making them non-informative for the decoder. Our distance measure, the correlation distance, while it enforces the representations to correlate, it does not constrain the scale of the values in the contextual vectors. This measure enforces the sentence distribution within the same language to be similar between all languages. However, since we are not constraining the scale, each language can be represented in a different space region.

Table \ref{tab:examples} shows some examples of English translations (using the English decoder) when using the three different encoders (English, Turkish and Kazakh) together with the reference translation. 

\begin{table*}[h!]
\begin{center}
\small
\begin{tabular}{|p{1.5cm}|p{12cm}|}
\hline
\textbf{System} & \textbf{Sentence}  \\ \hline
Reference       &    it was a terrific season.                \\
EN              &    we had a strong season .                \\
TK              &    it was a terrific season.\\
KK              &    we made a very big season .                \\ \hline
Reference       &    in London and Madrid it is completely natural for people with serious handicaps to be independently out in public, and they can use the toilets, go to the museum, or wherever ...               \\
EN              &    in London and Madrid , it is very normal for people with severe disability to be left to the public and be able to serve , to the museum , where ...
             \\
TK              &     in London and Madrid it is completely natural for people with serious handicaps to be independently out in public, and they can use the toilets, go to the museum, or wherever ...      \\
KK              &   in London and Madrid, it is quite common for people with a heavy disability to travel on their own in public spaces; they can go to the toilets, to the museum, anywhere ...       \\ \hline
Reference       &    from the Czech viewpoint, it seems they tend to put me to the left.    \\
EN              &    from a Czech point of view, I have the impression that people see me more than on the left.               \\
TK              &    from the Czech viewpoint, it seems they tend to put me to the left.   \\
KK              &    from a Czech point of view , I have the impression that people are putting me on the left .     \\ \hline
\end{tabular}
\caption{\label{tab:examples} English AutoEncoder (EN) and Translation examples from Turkish (TK), Kazakh (KK) to English compared to the Reference.}
\end{center}
\end{table*}

\section{Conclusions}
\label{conclusions}

We proposed a novel translation architecture which aims at a common intermediate representation for the benefit of incremental training in machine translation.

While there are already some machine translation systems where the implicit emergence of an internal interlingua representation is suggested, the main proposed difference in the current paper is forcing the Neural Machine Translation system to learn an intermediate multilingual representation while using independent encoders and decoders. This is achieved by combining the maximum likelihood loss, normally used in Neural Machine Translation, together with an extra loss term that computes a measure of the distance between intermediate representations in different languages. 

By achieving an interlingua representation, encoders and decoders are decoupled by having the interlingua as an interface. This leads to enabling every possible combination of encoder and decoder, effectively turning the quadratic needs, in terms of training data and resources, to linear. Furthermore, such a decoupling also allows training encoders/decoders to/from a new language that only has parallel data with one of the already supported languages, enabling the translation to/from any of the other supported languages. As a consequence, our architecture allows to extend the bilingual system to a multilingual system by incremental training. This particularity allows for quickly deploying new systems when requiring new languages into the system, which is highly relevant in scenarios like humanitarian crisis. 

 We have evaluated our model on different low-resourced language pairs. Our model outperforms current bilingual systems in the low-resourced setting (but not in the high-resourced) in addition to presenting a flexible architecture that enables scaling to new languages (achieving multilingual and zero-shot translation) without retraining languages in the system. Our approach supersedes pivoting approaches again in the context of a low-resourced setting. One of the next steps will be to exploit monolingual data in our architecture further avoiding dependency on the availability of parallel data.


\begin{acknowledgments}
This work is supported in part by the Google Faculty Research Award 2018, Spanish Ministerio de Econom\'ia y Competitividad, the European Regional  Development  Fund  and  the  Agencia  Estatal  de  Investigaci\'on,  through  the  postdoctoral  senior grant Ram\'on y Cajal, the contract TEC2015-69266-P (MINECO/FEDER,EU) and the contract PCIN-2017-079 (AEI/MINECO).
\end{acknowledgments}

\bibliography{relwork.bib}
\bibliographystyle{apa}

\end{document}